\documentclass{article}

\usepackage{arxiv}
\usepackage[T1]{fontenc}
\usepackage[utf8]{inputenc}
\usepackage{graphicx}
\usepackage{booktabs}
\usepackage{multirow}
\usepackage{amsmath,amssymb}
\usepackage{rotating}
\usepackage{tabularx}
\usepackage{microtype}
\usepackage{tikz}
\usetikzlibrary{trees}
\usepackage{hyperref}
\usepackage{orcidlink}
\usepackage[nameinlink,capitalise,noabbrev]{cleveref}

\providecommand{\doi}[1]{\href{https://doi.org/\detokenize{#1}}{https://doi.org/\detokenize{#1}}}

\graphicspath{{fig/}}

\title{Multi-Class vs.\ Multi-Label BERT for CVE-to-CWE Mapping:\\How Taxonomy Structure Shapes the Errors\thanks{Accepted for publication at ICANN 2026 (International Conference on Artificial Neural Networks). This is the authors' version of the work; the final authenticated version will be published by Springer.}}
\renewcommand{\shorttitle}{MC vs.\ ML BERT for CVE-to-CWE Mapping}
\date{}

\author{%
  Ana Schwengber Kelm\,\orcidlink{0009-0006-4177-8638} \\
  mindsquare AG\\
  Bielefeld, Germany\\
  \texttt{ana.luisa.kelm@mindsquare.de}
  \And
  Christian Bockermann\,\orcidlink{0000-0001-6570-0583} \\
  Bochum University of Applied Sciences\\
  Bochum, Germany\\
  \texttt{christian.bockermann@hs-bochum.de}
  \And
  J\"org Frochte\,\orcidlink{0000-0002-5908-5649} \\
  Bochum University of Applied Sciences\\
  Bochum, Germany\\
  \texttt{joerg.frochte@hs-bochum.de}
}

\begin{document}
\maketitle

\begin{abstract}
Assigning Common Weakness Enumeration (CWE) categories to Common Vulnerabilities and Exposures (CVE) records remains an important but largely manual step in vulnerability analysis. We study this task as a text classification problem and compare two modelling choices: a \emph{multi-class} formulation that predicts a single CWE per CVE and a \emph{multi-label} formulation that allows multiple assignments. Three transformer encoders (BERT Base, SecureBERT, and CySecBERT) are evaluated on three nested label spaces (83, 47, and 25 classes).

Multi-class training achieves higher macro-F1 across all settings, although the gap to multi-label narrows from 21 to 2 percentage points as the label space shrinks. Post-hoc threshold optimisation on the multi-label side closes this gap on the 25-class setting. Confusion analysis shows that the dominant misclassification patterns follow the CWE hierarchy and are shared across all three encoders (Pearson $r > 0.92$), which suggests that the error structure is driven more by taxonomy design than by encoder choice. A hierarchy-relaxed evaluation that forgives within-family confusions raises macro-F1 from ${\sim}$81\% to ${\sim}$90\%, indicating that strict metrics understate branch-level classifier quality. CySecBERT achieves the strongest results overall, with statistically significant gains concentrated in the multi-label setting.
\end{abstract}


\section{Introduction}
\label{sec:introduction}

In today's increasingly interconnected digital landscape, modern IT infrastructures are becoming
frequent targets of cyberattacks, many of which are only made possible due to flaws and vulnerabilities
in the underlying software. To address these threats systematically, the {\em Common Vulnerabilities
and Exposures} (CVE) program provides a standardized mechanism for identifying and cataloging publicly
disclosed vulnerabilities~\cite{cveOverview}. This system enables interoperable referencing of security-relevant
software defects across tools and organizations.

Identifying individual vulnerabilities is, however, only the first step: effective
vulnerability management requires understanding root causes. This is where the {\em Common Weakness
Enumeration} (CWE) becomes essential~\cite{mitre_cwe_guidance}. While a CVE describes a specific
incident, CWE provides a community-developed taxonomy of the underlying weakness types.
Mapping CVEs to CWEs enables pattern recognition across incidents~\cite{nistCVEsProcess},
systematic mitigation~\cite{howard2009improving,dodiya2021trend}, and aggregation at the
level of root causes~\cite{aghaei2023cvedrill}.

This mapping is currently performed largely by hand, and the workload is growing rapidly:
annual CVE volume has risen from approximately 1{,}500 in 1999 to over 40{,}000 in
2024~\cite{deepstrike2025vuln}. Automated classification is therefore not merely
convenient but operationally necessary.

From a machine-learning perspective, the task raises a natural modelling question.
A CVE record is typically assigned to one CWE class, but a vulnerability may exhibit
multiple weakness facets, making the task conceptually multi-label. At the same time,
the class distribution is heavily long-tailed, and CWE annotations mix different
abstraction levels, ranging from broad ``Pillar'' categories to specific ``Variant'' nodes.
Understanding how these factors interact with the choice of output formulation is
the central question of this paper.

We focus on three research questions:
\emph{(i)}~how does the choice between multi-class and multi-label formulation affect
performance across label-space sizes?
\emph{(ii)}~does domain-adaptive pretraining improve results over a general-purpose encoder?
\emph{(iii)}~what role do the decision threshold and the CWE hierarchy structure play
in interpreting classifier quality?

\paragraph{Contributions.}
First, we compare multi-class and multi-label formulations under closely matched settings and show that multi-class achieves higher macro-F1 across all slices. Second, we show that the standard sigmoid threshold of~0.5 is suboptimal for the multi-label setting and that threshold tuning substantially reduces the gap to multi-class. Third, we show that the dominant errors follow the CWE hierarchy, are shared across all three encoders, and become much less severe under a hierarchy-relaxed evaluation that reaches ${\sim}$90\% macro-F1 on the 25-class slice.

\section{Background}
\label{sec:background}

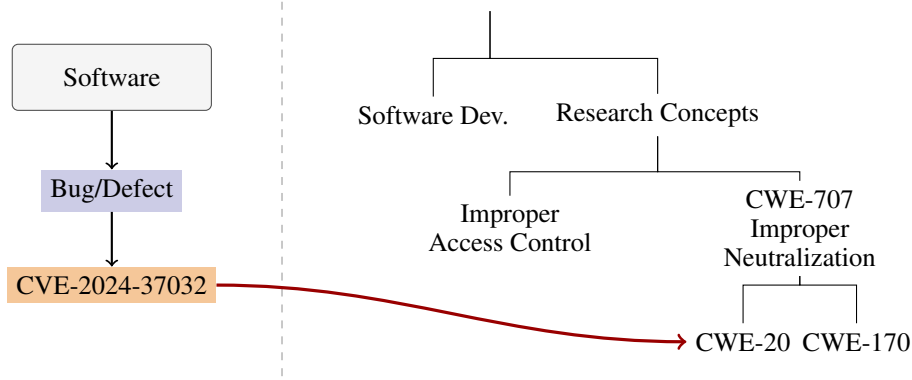
\begin{figure}[t]
\centering
\begin{tikzpicture}
\begin{scope}[shift={(-1,0)}]
\node[fill=black!4,draw=black!70,inner sep=2ex,align=center,text width=2cm,rectangle,rounded corners=.5ex] (S) at (0,0) {Software};
\node[fill=blue!50!black!20,rectangle] (B) at (0,-1.5) {Bug/Defect};
\draw[->,thick] (S.south) -- (B.north);
\node[fill=orange!90!black!40,rectangle] (C) at (0,-2.75) {CVE-2024-37032};
\draw[->,thick] (B.south) -- (C.north);
\end{scope}
\draw[dashed,black!40] (1.25,1) -- (1.25,-4);
  \node at (4,1) {}
    [edge from parent fork down]
    child {node {Software Dev.}}
    child {node[anchor=west] {Research Concepts}
      child[child anchor=north] {node[anchor=east,align=center] {Improper\\ Access Control}}
      child {node[anchor=west, align=center] {CWE-707 \\ Improper\\ Neutralization}}{
         child {node[rectangle] (CWE) {CWE-20}}
         child {node[rectangle] {CWE-170}}
      }
    };
\draw[->,very thick,red!60!black] (C) to[out=0,in=180] (CWE);
\end{tikzpicture}
\caption{\label{fig:cwe_tree}A CVE mapped to a CWE taxonomy node.
The hierarchy mixes abstraction levels: CWE-707 (``Pillar'') and CWE-20 (``Base'')
coexist as valid targets, creating uneven supervision for classifiers.}
\end{figure}

\paragraph{CVE, CWE, and annotation granularity.}
A CVE record describes a specific publicly known vulnerability~\cite{cveOverview}.
CWE is a community-developed taxonomy of weakness types that can lead to
vulnerabilities~\cite{cweAbout}. \Cref{fig:cwe_tree} illustrates the relationship:
a concrete vulnerability (CVE-2024-37032) is mapped to a specific node (CWE-20)
in the CWE hierarchy.

Critically, CWE contains abstraction levels that differ substantially in specificity.
MITRE's mapping guidance warns that overly abstract classes are less useful for
root-cause analysis~\cite{mitre_cwe_guidance}, yet abstract nodes are still widely
assigned in practice. In our data, CWE-707 (a ``Pillar'') appears 194~times,
while its more specific descendants occur in single-digit counts. For a text classifier,
the training signal is therefore not only sparse but also inconsistent in granularity, a point we return to when analysing confusion patterns.

\paragraph{Transformer models and domain adaptation.}
BERT~\cite{bertog} established bidirectional encoder pretraining for downstream
classification; RoBERTa~\cite{liu2019roberta} refined the recipe. For specialised
domains, continuing pretraining on in-domain corpora (DAPT) has proven
effective~\cite{gururangan2020dapt}. SecureBERT~\cite{aghaei2022securebert}
and CySecBERT~\cite{bayer2022cysecbert} apply this idea to cybersecurity text.

\section{Related Work}
\label{sec:related}

Early work on vulnerability text classification relied on classical methods with
bag-of-words features~\cite{na2017naivebayes}. ThreatZoom~\cite{threatZoomAghaei2020}
introduced a hierarchy-aware neural network using TF-IDF features; more recent
approaches combine pretrained language models with hierarchical
classification~\cite{aghaei2023cvedrill}, enhanced
attention~\cite{suwu2024refiningcvetocwe}, or multi-objective
training~\cite{tiwari2025advancing}.

Simonetto et al.~\cite{simonetto_etal_2025} take a data-centric perspective:
they use LLMs to extract key terms from CVE descriptions and show that embedding
these distilled representations improves classification by up to 8.88~percentage
points. Notably, they also find that direct LLM prompting (chain-of-thought and
zero-shot) underperforms the supervised approach, an observation that supports
the use of fine-tuned encoder models. Such encoders also have practical deployment
advantages: fewer than 125M parameters, trainable in minutes on a single GPU, and
deployable in air-gapped security environments.

These studies differ in class sets, hierarchy levels, and evaluation protocols, which makes direct comparison difficult. More importantly, prior work typically commits to either a multi-class or a multi-label formulation instead of comparing the two directly. It also leaves largely open whether the resulting errors are model-specific or induced by the CWE taxonomy itself.
Our work is complementary to these efforts: rather than proposing a new model, it addresses both questions by providing an empirical analysis which looks at the impact of the output formulation itself.  By comparing MC and ML under closely matched settings, we further investigate the role of threshold behavior and the structure of the remaining taxonomy-driven errors in shaping the performance of CVE-to-CWE mapping.

\section{Methodology}
\label{sec:method}

\paragraph{Task formulation.}
Given a CVE description~$x$ and a set of CWE classes~$\mathcal{C}$, the
\emph{multi-class} (MC) setting predicts one class $y \in \mathcal{C}$ via softmax.
The \emph{multi-label} (ML) setting predicts a subset $Y \subseteq \mathcal{C}$ through
independent per-class sigmoid outputs, binarised at a threshold~$\tau$ (default~0.5).

\paragraph{Dataset.}
Our experiments build on CVE records maintained by the MITRE
Corporation~\cite{cveOverview}, the de-facto standard for publicly disclosed
vulnerabilities (snapshot accessed April 2025).
The resulting dataset covers 249{,}867~records (1999--2024). The annotation landscape is
sparse: 77\% carry no CWE assignment, 22\% have exactly one, and fewer than 1\%
have multiple CWE labels. This overwhelming single-label bias means that multi-label
supervision, while conceptually appropriate, rests on very limited ground truth.

Of the 968~CWE types in the taxonomy, only 664 appear in the data, with an extremely
skewed frequency distribution. After filtering to classes with $\geq$100~linked CVE
records, we obtain 83~classes (the count is determined by the frequency threshold,
not a target; analogous thresholds of $\geq$205 and $\geq$570 yield 47 and 25~classes).
From this set we derive three nested label
spaces (83, 47, and 25~classes), ordered by descending frequency. This allows us to
study how label-space size affects the MC--ML trade-off under otherwise constant conditions.

For MC, each record is mapped to its first listed CWE; for ML, all labels are retained. This follows the annotation order provided in the CVE record and treats the first label as the primary target in the single-label setting.
An important methodological caveat: because MC applies targeted undersampling of the four
most frequent classes (capped at 2{,}000 instances), the MC and ML training sets are not
identical. Comparisons therefore reflect both the formulation difference and a data-composition
effect. We return to this point in the discussion.

\paragraph{Models.}
We compare three encoders:
\textbf{BERT Base}~\cite{bertog} (general baseline),
\textbf{SecureBERT}~\cite{aghaei2022securebert} (RoBERTa-family, security-pretrained),
and \textbf{CySecBERT}~\cite{bayer2022cysecbert} (BERT-family, security-pretrained).
All models are \emph{fully} fine-tuned, updating all ${\sim}$110M parameters including encoder
layers, using \texttt{torch.optim.AdamW} with learning rate $2 {\times} 10^{-5}$,
weight decay~0.05, batch size~8, and 6~epochs. Each configuration is repeated across
five random seeds with an 80/20 stratified split. Class-weighted loss is used throughout.

\paragraph{Imbalance mitigation and optimiser interaction.}
\Cref{tab:optim} shows that the effect of explicit balancing is tied to the specific optimization setting used in our experiments, without specific parameter-tuning. Under AdamW, class weighting and undersampling change macro-F1 by less than 1~pp and do not produce dead classes. Under SGD with the same learning-rate schedule, the same task becomes unstable: without mitigation, training collapses almost completely (81 of 83~classes receive no predictions). This pattern is consistent with recent work suggesting that Adam-style per-parameter normalisation partially offsets heterogeneous class frequencies~\cite{kunstner2024heavytailed}. Since AdamW is standard in transformer fine-tuning, the practical implication is that explicit balancing may matter less than often assumed when the optimiser already provides substantial robustness to skewed gradients.

\begin{table}[t]
\centering
\caption{Effect of imbalance mitigation under different optimisers (CySecBERT, MC,
83~classes, single seed). ``Dead'' = classes with F1${\approx}$0.}
\label{tab:optim}
\setlength{\tabcolsep}{4pt}
\small
\begin{tabular}{llcc}
\toprule
\textbf{Optimiser} & \textbf{Mitigation} & \textbf{Macro-F1} & \textbf{Dead} \\
\midrule
AdamW & weighted + undersampled & 66.9\% & 0 \\
AdamW & none                    & 66.7\% & 0 \\
SGD   & weighted + undersampled & 48.4\% & 9 \\
SGD   & none                    &  0.4\% & 81 \\
\bottomrule
\end{tabular}
\end{table}

\paragraph{Evaluation.}
We report macro-F1 as the primary metric because it weights all classes equally and is
therefore most sensitive to minority-class performance. Micro-F1, weighted-F1, and
(for ML) samples-F1 are reported for completeness. In addition, we introduce a
\emph{hierarchy-relaxed macro-F1}: a misclassification is counted as correct if
the true and predicted CWE belong to the same family branch. The family groupings
are defined manually based on the CWE hierarchy (e.g., Memory/Buffer: CWE-119, 120,
121, 122, 125, 787; Injection: CWE-20, 77, 78, 79, 89, 94; Access Control: CWE-269,
284, 287, 862). This remains a heuristic operationalisation, and alternative groupings would yield different values, but it offers a useful estimate of how often the classifier reaches the correct region of the taxonomy even when it misses the exact CWE node.

\section{Results}
\label{sec:results}

We organise the results around three questions: which formulation performs better
(MC vs.\ ML), whether the gap can be closed through threshold tuning, and what the
remaining errors reveal about the CWE taxonomy.

\begin{figure*}[t]
\centering
\includegraphics[width=0.95\linewidth]{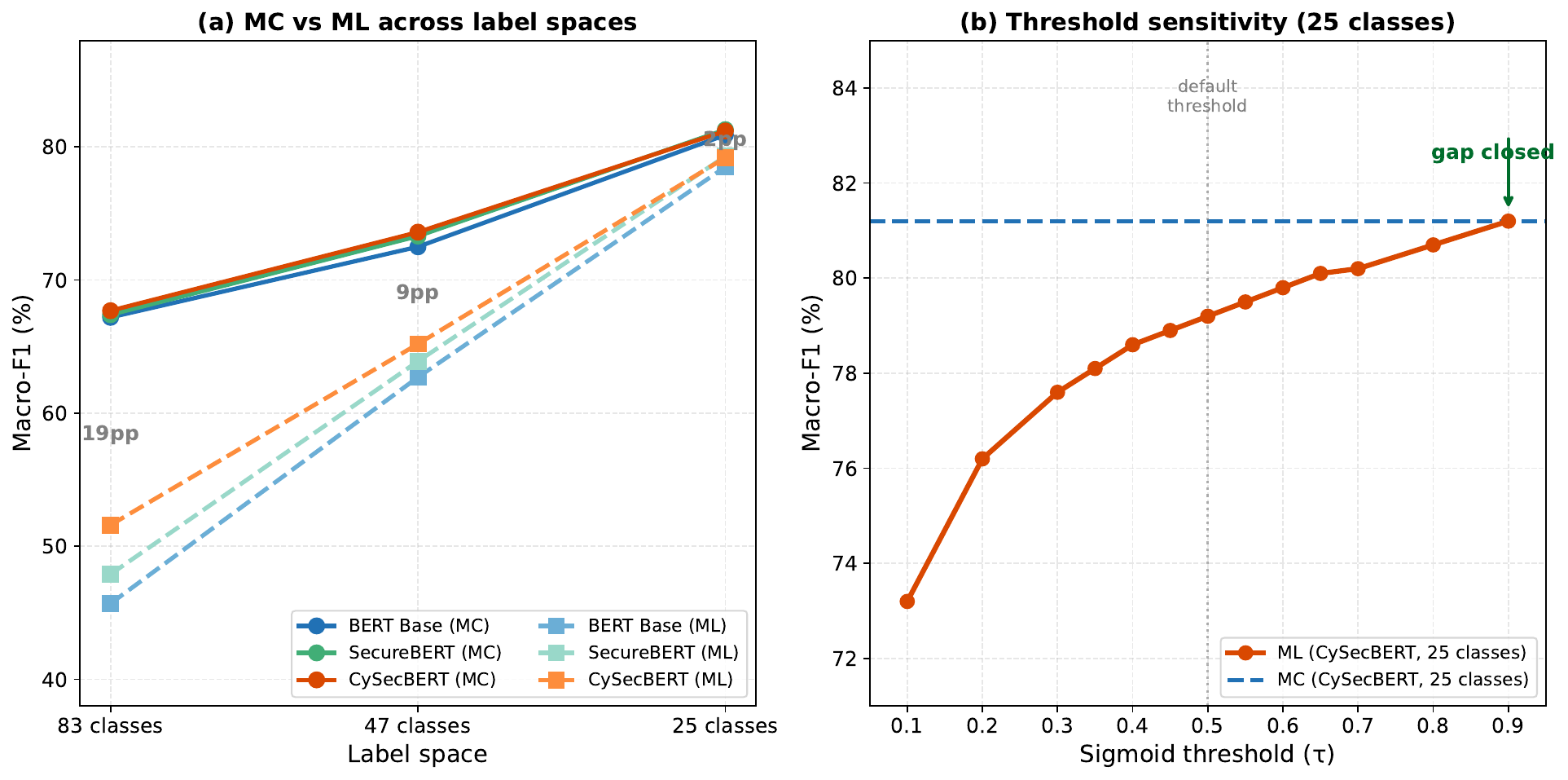}
\caption{(a)~Macro-F1 across label spaces: MC (solid) outperforms ML (dashed)
throughout; the gap (annotated) narrows from 21~pp to 2~pp.
(b)~Threshold sensitivity for CySecBERT ML on 25~classes: raising $\tau$ from the
default~0.5 to~0.9 closes the gap to MC.}
\label{fig:combined}
\end{figure*}

\paragraph{Multi-class versus multi-label.}
\Cref{tab:mc_results,tab:ml_results} present the full results; the central
comparison is distilled in \Cref{tab:comparison} and visualised in
\Cref{fig:combined}(a).

\begin{table*}[t]
\centering
\caption{Multi-class results: mean $\pm$ std F1 (\%) over five seeds. Best macro-F1 per column in bold.}
\label{tab:mc_results}
\setlength{\tabcolsep}{5pt}
\begin{tabular}{llccc}
\toprule
& & \textbf{83 classes} & \textbf{47 classes} & \textbf{25 classes} \\
\midrule
\multirow{3}{*}{\footnotesize BERT Base}
  & Macro-F1    & $67.2 \pm 0.6$ & $72.5 \pm 0.2$ & $80.9 \pm 0.3$ \\
  & Micro-F1    & $73.7 \pm 0.3$ & $77.5 \pm 0.2$ & $82.9 \pm 0.2$ \\
  & Weighted-F1 & $73.8 \pm 0.3$ & $77.5 \pm 0.2$ & $82.8 \pm 0.2$ \\
\midrule
\multirow{3}{*}{\footnotesize SecureBERT}
  & Macro-F1    & $67.4 \pm 0.4$ & $73.3 \pm 0.2$ & $81.3 \pm 0.3$ \\
  & Micro-F1    & $73.9 \pm 0.2$ & $78.1 \pm 0.2$ & $83.3 \pm 0.3$ \\
  & Weighted-F1 & $74.0 \pm 0.2$ & $78.1 \pm 0.2$ & $83.2 \pm 0.3$ \\
\midrule
\multirow{3}{*}{\footnotesize CySecBERT}
  & Macro-F1    & $\mathbf{67.7 \pm 0.5}$ & $\mathbf{73.6 \pm 0.3}$ & $\mathbf{81.2 \pm 0.3}$ \\
  & Micro-F1    & $74.4 \pm 0.4$ & $78.4 \pm 0.2$ & $83.2 \pm 0.3$ \\
  & Weighted-F1 & $74.4 \pm 0.4$ & $78.3 \pm 0.2$ & $83.1 \pm 0.3$ \\
\bottomrule
\end{tabular}
\end{table*}

\begin{table*}[t]
\centering
\caption{Multi-label results: mean $\pm$ std F1 (\%) over five seeds at $\tau{=}0.5$. Best macro-F1 per column in bold.}
\label{tab:ml_results}
\setlength{\tabcolsep}{5pt}
\begin{tabular}{llccc}
\toprule
& & \textbf{83 classes} & \textbf{47 classes} & \textbf{25 classes} \\
\midrule
\multirow{4}{*}{\footnotesize BERT Base}
  & Macro-F1    & $45.7 \pm 0.3$ & $62.7 \pm 0.3$ & $78.5 \pm 0.3$ \\
  & Samples-F1  & $71.8 \pm 0.1$ & $79.7 \pm 0.2$ & $87.2 \pm 0.2$ \\
  & Micro-F1    & $56.9 \pm 0.4$ & $72.2 \pm 0.3$ & $84.5 \pm 0.2$ \\
  & Weighted-F1 & $65.9 \pm 0.2$ & $75.8 \pm 0.1$ & $85.1 \pm 0.2$ \\
\midrule
\multirow{4}{*}{\footnotesize SecureBERT}
  & Macro-F1    & $47.9 \pm 0.6$ & $63.9 \pm 0.4$ & $79.3 \pm 0.2$ \\
  & Samples-F1  & $73.2 \pm 0.2$ & $80.5 \pm 0.2$ & $87.9 \pm 0.1$ \\
  & Micro-F1    & $59.1 \pm 0.5$ & $72.9 \pm 0.4$ & $85.2 \pm 0.1$ \\
  & Weighted-F1 & $67.5 \pm 0.3$ & $76.4 \pm 0.2$ & $85.7 \pm 0.1$ \\
\midrule
\multirow{4}{*}{\footnotesize CySecBERT}
  & Macro-F1    & $\mathbf{51.6 \pm 0.4}$ & $\mathbf{65.2 \pm 0.3}$ & $\mathbf{79.2 \pm 0.4}$ \\
  & Samples-F1  & $75.0 \pm 0.3$ & $81.0 \pm 0.1$ & $87.6 \pm 0.1$ \\
  & Micro-F1    & $62.7 \pm 0.4$ & $74.5 \pm 0.2$ & $85.2 \pm 0.2$ \\
  & Weighted-F1 & $69.5 \pm 0.2$ & $77.2 \pm 0.1$ & $85.7 \pm 0.2$ \\
\bottomrule
\end{tabular}
\end{table*}

\begin{table}[t]
\centering
\caption{Macro-F1 (\%) across formulations and label spaces.}
\label{tab:comparison}
\setlength{\tabcolsep}{3.5pt}
\begin{tabular}{l|ccc|ccc}
\toprule
 & \multicolumn{3}{c|}{\textbf{Multi-class}} & \multicolumn{3}{c}{\textbf{Multi-label}} \\
 & 83 & 47 & 25 & 83 & 47 & 25 \\
\midrule
BERT Base  & 67.2 & 72.5 & 80.9 & 45.7 & 62.7 & 78.5 \\
SecureBERT & 67.4 & 73.3 & 81.3 & 47.9 & 63.9 & 79.3 \\
CySecBERT  & 67.7 & 73.6 & 81.2 & 51.6 & 65.2 & 79.2 \\
\bottomrule
\end{tabular}
\end{table}

CySecBERT achieves strong results in every setting,
confirming the value of
domain-adaptive pretraining (differences to BERT Base are 0.3--5.9~pp and consistent
across seeds). MC outperforms ML on macro-F1 across all label spaces and all models
(\Cref{fig:combined}a). The gap narrows systematically, from ${\sim}$21~pp on
83~classes to ${\sim}$2~pp on 25~classes, but no crossover appears.

Because MC uses undersampled training data whereas ML does not, one possible concern is that the observed gap reflects data composition rather than formulation choice. We therefore also train MC \emph{without} undersampling, i.e.\ on the same data basis as ML. The result is nearly unchanged: CySecBERT MC without undersampling achieves
$66.7 {\pm} 0.3$ (vs.\ $51.6 {\pm} 0.4$ ML),
$72.6 {\pm} 0.4$ (vs.\ $65.2 {\pm} 0.3$), and
$81.5 {\pm} 0.4$ (vs.\ $79.2 {\pm} 0.4$) on
83, 47, and 25~classes.
The remaining gaps of 15, 7, and 2~pp suggest that the MC advantage is primarily a formulation effect rather than a data-composition artefact. This hints that the performance gain is an inherent advantage of 
the MC model approach.

The narrowing suggests that ML is limited mainly by sparse per-class supervision
in the larger label spaces, rather than by an inherent weakness of the formulation.

\paragraph{Threshold sensitivity and gap closure.}
All ML results above use the conventional sigmoid threshold $\tau{=}0.5$. Is this
default actually optimal? Figure \Cref{fig:combined}(b) shows the results of an internal
threshold validation run for CySecBERT on 25~classes: macro-F1 increases \emph{monotonically} with $\tau$, rising from
77.6\% at $\tau{=}0.3$ to \textbf{81.2\%} at $\tau{=}0.9$, matching the MC
result. The default leaves ${\sim}$2~pp on the table.

To test whether this result is merely a test-set artefact, we additionally run a leave-one-seed-out validation. For each of the five seeds, we select $\tau$ on the remaining four and evaluate on the held-out seed.
In this setting, the gap closes at 0.90 in every fold,
and the mean validated macro-F1 remains $81.2 \pm 0.2$, matching the post-hoc estimate. This supports the interpretation that ML systematically overpredicts at moderate confidence levels and benefits from a stricter decision boundary.

\paragraph{Residual error structure.}
Once MC and ML become nearly indistinguishable on 25~classes, the next question is not the error rate alone but the \emph{structure} of the remaining errors. Confusion matrices reveal a striking regularity.
To compare error patterns across models, we row-normalise each confusion matrix,
set the diagonal to zero, and compute the Pearson correlation between the flattened
off-diagonal vectors for each model pair. All correlations exceed $r = 0.92$ on every
label space: the three encoders make \emph{quantitatively different} but
\emph{structurally identical} errors.

\begin{table}[t]
\centering
\caption{Top-5 misclassification pairs (MC, 25~classes, consistent across models).
All correspond to hierarchically related CWEs.}
\label{tab:confusions}
\setlength{\tabcolsep}{3pt}
\small
\begin{tabular}{llcl}
\toprule
\textbf{True} & \textbf{Predicted} & \textbf{Rate} & \textbf{Relationship} \\
\midrule
CWE-77  & CWE-78  & 28--30\% & Parent $\to$ Child \\
CWE-269 & CWE-284 & 16--17\% & Sibling \\
CWE-287 & CWE-284 & 9--10\%  & Sibling \\
CWE-78  & CWE-77  & 8--9\%   & Child $\to$ Parent \\
CWE-476 & CWE-20  & 8--9\%   & Specific $\to$ Abstract \\
\bottomrule
\end{tabular}
\end{table}

\begin{figure}[t]
\centering
\includegraphics[width=0.95\linewidth]{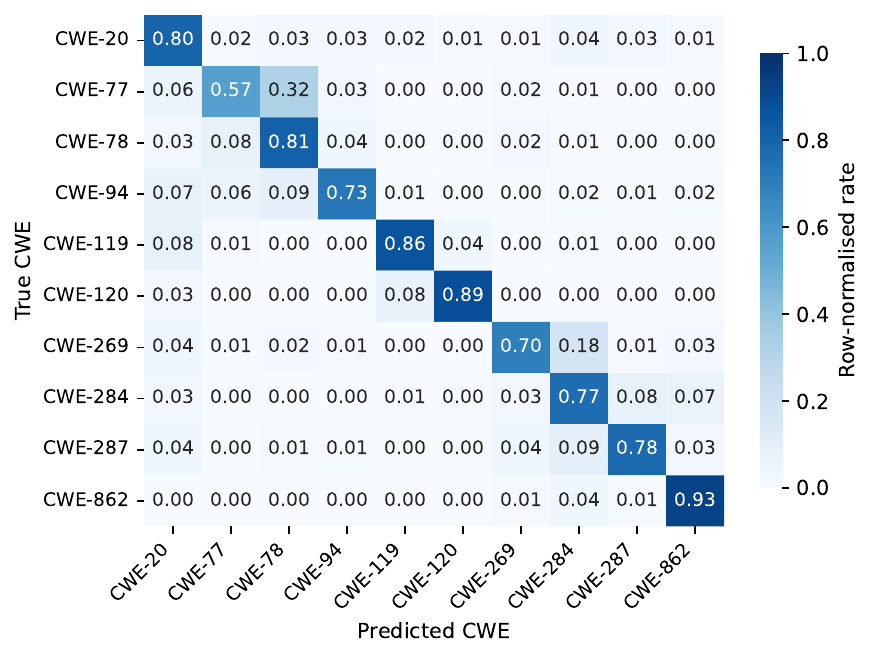}
\caption{Zoomed confusion matrix for CySecBERT (MC, 25~classes, averaged over 5~seeds),
showing the 10~most confusion-prone classes. Off-diagonal mass concentrates at
hierarchy-related pairs (CWE-77/78, CWE-269/284/287).}
\label{fig:confusion}
\end{figure}

\Cref{tab:confusions} lists the dominant pairs, all of which are parent--child or sibling
relationships in the CWE tree. \Cref{fig:confusion} visualises the pattern for
CySecBERT: a strong diagonal with small off-diagonal clusters at exactly the
hierarchy-related positions.

\paragraph{Quantifying the impact: hierarchy-relaxed evaluation.}
The hierarchy-relaxed macro-F1 (defined in Section~\ref{sec:method}) captures how
often the classifier at least identifies the correct weakness \emph{family}, even
when the exact CWE is wrong. \Cref{tab:relaxed} shows the results across all three
label spaces.

\begin{table}[t]
\centering
\caption{Composition of the label space and its effect on hierarchy-relaxed
evaluation (CySecBERT MC; BERT Base and SecureBERT within $\pm$0.5~pp).
``Family coverage'' is the share of classes that have $\geq$1 family member
in the label space; ``within-family errors'' is the share of all misclassifications
that land in the same CWE family.}
\label{tab:relaxed}
\setlength{\tabcolsep}{4pt}
\small
\begin{tabular}{lcccccc}
\toprule
& \textbf{Family} & \textbf{Within-fam.} & \multicolumn{2}{c}{\textbf{Macro-F1}} & \\
\textbf{Slice} & \textbf{coverage} & \textbf{errors} & \textbf{Strict} & \textbf{Relaxed} & \textbf{Gain} \\
\midrule
83 cl. &  19\% &  17\% & 67.7 & 69.8 & +2.1 \\
47 cl. &  36\% &  23\% & 73.6 & 77.5 & +3.9 \\
25 cl. &  72\% &  42\% & 81.2 & 89.8 & +8.7 \\
\bottomrule
\end{tabular}
\end{table}

\Cref{tab:relaxed} makes the gain pattern interpretable. On 25~classes, 72\% of classes have at least one family member in the label space, so within-family confusions are both possible and frequent (42\% of all errors). The relaxed metric can therefore credit many of these near-misses, which raises macro-F1 by nearly 9~pp. On 83~classes, by contrast, only 19\% of classes have family members in the slice. The 58 additional classes are taxonomic singletons, so their errors are necessarily cross-family and cannot benefit from the relaxed criterion.

The drop in relaxation gain from 25 to 83~classes is therefore a composition effect of the label space rather than evidence of lower classifier quality. In larger slices, strict macro-F1 increasingly conflates genuine cross-branch mistakes with taxonomy-near misses that arise only because many classes lack family neighbours in the evaluation space. This matters especially for vulnerability triage: operational workflows often care more about routing a CVE into the correct weakness family than about distinguishing between adjacent nodes such as CWE-77 and CWE-78 on the first pass.

\paragraph{Summary: how evaluation changes the conclusion.}
\Cref{tab:bridge} condenses the central message for CySecBERT on 25~classes. Under the default evaluation setup, MC leads ML by 2~pp. After threshold tuning, the gap disappears. Under hierarchy-relaxed evaluation, both formulations reach ${\sim}$90\%, which suggests that the practical difference between MC and ML becomes small once the taxonomy structure is taken into account explicitly.

\begin{table}[t]
\centering
\caption{CySecBERT on 25~classes: how the apparent MC--ML comparison changes under three increasingly permissive evaluation views. ``Standard'' uses the default ML threshold $\tau{=}0.5$; ``threshold-tuned'' uses the validated optimum $\tau{=}0.9$; ``hierarchy-relaxed'' forgives within-family confusions. All values are macro-F1 (\%).}
\label{tab:bridge}
\setlength{\tabcolsep}{5pt}
\begin{tabular}{lcc}
\toprule
\textbf{Evaluation setup} & \textbf{MC} & \textbf{ML} \\
\midrule
Standard ($\tau{=}0.5$) & 81.2 & 79.2 \\
Threshold-tuned ($\tau{=}0.9$) & 81.2 & 81.2 \\
Hierarchy-relaxed & 89.8 & ${\sim}$90 \\
\bottomrule
\end{tabular}
\end{table}

\section{Discussion}
\label{sec:discussion}

\paragraph{Implications for vulnerability management pipelines.}
For operational CVE-to-CWE assignment, MC is the simpler choice: it yields higher
macro-F1 throughout (even on matched training data), requires no threshold tuning,
and produces a single prediction. ML becomes competitive only with threshold tuning
and may become more relevant as annotation practices evolve toward richer multi-label
ground truth. In either case, the hierarchy-relaxed evaluation suggests that deployed
classifiers are substantially more useful than standard macro-F1 indicates: at
branch level, accuracy exceeds 90\% on the most frequent 25~classes. For automated
triage pipelines, where identifying the correct weakness \emph{category} matters
more than the exact CWE identifier, this is the more operationally relevant measure.

\paragraph{Error structure as a diagnostic tool.}
The near-identical confusion patterns across all three encoders ($r > 0.92$) suggest
that the dominant errors are driven by taxonomy ambiguity, since parent--child pairs like
CWE-77/78 are textually very similar, rather than by model limitations. This has a
practical implication: improving classification on these pairs likely requires better
annotation guidelines or hierarchy-aware training objectives, not a better encoder.

\paragraph{Does domain adaptation help?}
CySecBERT achieves the numerically highest macro-F1 across all settings.
However, paired $t$-tests over the five seeds reveal that the advantage is
statistically significant only in certain configurations: on the ML side,
CySecBERT significantly outperforms BERT Base on all three label spaces
($p < 0.02$) and SecureBERT on 83 and 47~classes ($p < 0.01$). On the MC side,
differences between the three encoders are smaller and mostly not significant
at $\alpha{=}0.05$; CySecBERT vs.\ BERT Base reaches significance only on
47~classes ($p{=}0.007$). This is expected: full fine-tuning on tens of thousands
of domain texts already adapts a general-purpose encoder substantially, leaving
less room for DAPT to add value. The practical conclusion is that domain-adaptive
pretraining provides a reliable advantage when fine-tuning data is limited or
training is short, but the benefit diminishes under aggressive full fine-tuning.

\paragraph{Limitations.}
Since our work focuses on conducting a controlled comparison of encoder behavior under
matched conditions, we do not investigate hierarchy-aware architectures or
employ classifier chains. For a real-world deployment, evaluation would focus much
more on domain-specific label correlations, demanding an extensive exploration of
different encoder settings.

By filtering to CWE classes with $\geq$100 records, we exclude the long tail that
matters in full deployment. Absolute scores are snapshot-specific; comparative trends
should be more robust. The threshold analysis is post-hoc on the test set.
The hierarchy-relaxed metric uses manually defined family groupings. A sensitivity
analysis with four grouping variants (from minimal parent--child pairs to broader
merged families) yields relaxed macro-F1 between 84\% and 91\% on 25~classes,
confirming that the finding is robust to the specific grouping; an automated derivation
from the CWE ontology would nevertheless be more principled.
Although a matched-data ablation confirms the MC advantage
(Section~\ref{sec:results}), the primary MC results use undersampled training data,
which limits strict score comparability with ML.

\paragraph{Future work.} 
Future work should prioritize the development of hierarchy-aware training 
objectives and metrics to address the taxonomic ambiguities driving current 
errors, while also investigating label correlation models to better capture 
dependencies between CWE categories. Regarding the long tail of infrequent 
classes, future research should assess whether the significant effort 
required to model these rare types is operationally justified for real-world 
triage, or if a hierarchy-relaxed approach -- identifying the correct weakness 
family instead of the exact node -- already provides sufficient practical 
utility for automated workflows.

\section{Conclusion}
\label{sec:conclusion}

We compared multi-class and multi-label learning for CVE-to-CWE mapping across three
Transformer encoders and three label-space sizes. Three findings stand out.
First, multi-class consistently achieves higher macro-F1; the gap narrows from 21
to 2~pp as the class set shrinks and can be narrowed further through threshold tuning
on the multi-label side. Second, the dominant misclassification patterns follow the
CWE hierarchy and are shared across all models ($r > 0.92$); a hierarchy-relaxed
evaluation raises macro-F1 to ${\sim}$90\%, suggesting that standard metrics
substantially underestimate how often the classifier identifies the correct weakness
family. Third, CySecBERT achieves the numerically best results throughout, with
statistically significant advantages primarily in the multi-label setting where
per-class supervision is sparser.

These findings point toward hierarchy-aware training objectives and evaluation metrics
as the most promising directions for future work on CVE-to-CWE mapping.

\bibliographystyle{splncs04}
\bibliography{references}

\begin{thebibliography}{10}
\providecommand{\url}[1]{\texttt{#1}}
\providecommand{\urlprefix}{URL }
\providecommand{\doi}[1]{https://doi.org/#1}

\bibitem{aghaei2023cvedrill}
Aghaei, E., Al-Shaer, E., Shadid, W., Niu, X.: Automated {CVE} analysis for
  threat prioritization and impact prediction (2023)

\bibitem{aghaei2022securebert}
Aghaei, E., Niu, X., Shadid, W., Al-Shaer, E.: {SecureBERT}: A domain-specific
  language model for cybersecurity (2022)

\bibitem{threatZoomAghaei2020}
Aghaei, E., Shadid, W., Al-Shaer, E.: {ThreatZoom}: Hierarchical Neural Network
  for {CVE}s to {CWE}s Classification, pp. 23--41. Springer (2020).
  \doi{10.1007/978-3-030-63086-7_2}

\bibitem{bayer2022cysecbert}
Bayer, M., Kuehn, P., Shanehsaz, R., Reuter, C.: {CySecBERT}: A domain-adapted
  language model for the cybersecurity domain (2022)

\bibitem{deepstrike2025vuln}
{DeepStrike}: Vulnerabilities statistics 2025: Record {CVE}s, zero-days \&
  exploit speed (oct 2025),
  \url{https://deepstrike.io/blog/vulnerability-statistics-2025}

\bibitem{bertog}
Devlin, J., Chang, M., Lee, K., Toutanova, K.: {BERT}: Pre-training of deep
  bidirectional transformers for language understanding. Proceedings of
  NAACL-HLT pp. 4171--4186 (2019)

\bibitem{dodiya2021trend}
Dodiya, B., Singh, U.K., Gupta, V.: Trend analysis of the cve classes across
  cvss metrics. Int. J. Comput. Appl  \textbf{183}(33),  23--30 (2021)

\bibitem{gururangan2020dapt}
Gururangan, S., Marasovi\'{c}, A., Swayamdipta, S., Lo, K., Beltagy, I.,
  Downey, D., Smith, N.A.: Don't stop pretraining: Adapt language models to
  domains and tasks (2020)

\bibitem{howard2009improving}
Howard, M.: Improving software security by eliminating the cwe top 25
  vulnerabilities. IEEE Security \& Privacy  \textbf{7}(3),  68--71 (2009)

\bibitem{kunstner2024heavytailed}
Kunstner, F., Yadav, R., Milligan, A., Schmidt, M., Bietti, A.: Heavy-tailed
  class imbalance and why adam outperforms gradient descent on language models.
  In: Advances in Neural Information Processing Systems. vol.~37 (2024)

\bibitem{liu2019roberta}
Liu, Y., Ott, M., Goyal, N., Du, J., Joshi, M., Chen, D., Levy, O., Lewis, M.,
  Zettlemoyer, L., Stoyanov, V.: {RoBERTa}: {A} robustly optimized {BERT}
  pretraining approach. CoRR  \textbf{abs/1907.11692} (2019),
  \url{http://arxiv.org/abs/1907.11692}

\bibitem{na2017naivebayes}
Na, S., Kim, T., Kim, H.: A study on the classification of common
  vulnerabilities and exposures using na\"{\i}ve bayes. In: Advances on
  Broad-Band Wireless Computing, Communication and Applications. pp. 657--662.
  Springer (2017)

\bibitem{nistCVEsProcess}
{National Institute of Standards and Technology}: {NVD} -- {CVE}s and the {NVD}
  process (2024), \url{https://nvd.nist.gov/general/cve-process}

\bibitem{simonetto_etal_2025}
Simonetto, S., Oosteven, R., \{Van Ede\}, T., Bosch, P., Jonker, W.: What
  matters most in vulnerabilities? key term extraction for cve-to-cwe mapping
  with llms. In: Kim, Y., Miyaji, A., Tibouchi, M. (eds.) Cryptology and
  Network Security. pp. 467--492. Lecture Notes in Computer Science, Springer,
  Germany (Nov 2026). \doi{10.1007/978-981-95-4434-9\_22}

\bibitem{suwu2024refiningcvetocwe}
Su, J., Wu, Y.: Refining {CVE}-to-{CWE} mapping with enhanced attention in
  {BERT}-based models. Applied and Computational Engineering  \textbf{71},
  107--112 (2024). \doi{10.54254/2755-2721/71/20241647}

\bibitem{cweAbout}
{The MITRE Corporation}: About {CWE} (2024),
  \url{https://cwe.mitre.org/about/index.html}

\bibitem{cveOverview}
{The MITRE Corporation}: {CVE} overview (2024),
  \url{https://www.cve.org/About/Overview}

\bibitem{mitre_cwe_guidance}
{The MITRE Corporation}: {CWE} usage guidance (2024),
  \url{https://cwe.mitre.org/documents/cwe_usage/guidance.html}

\bibitem{tiwari2025advancing}
Tiwari, H.: Advancing vulnerability classification with {BERT}: A
  multi-objective learning model (2025)

\end{thebibliography}

\end{document}